\pdfoutput=1

\documentclass[11pt]{article}

\usepackage[preprint]{acl}

\usepackage{times}
\usepackage{latexsym}
\usepackage{kotex}

\usepackage[T1]{fontenc}

\usepackage[utf8]{inputenc}

\usepackage{microtype}

\usepackage{inconsolata}

\usepackage{graphicx}
\usepackage{url}
\usepackage{amsmath,amssymb}
\usepackage{booktabs}
\usepackage{tabularx}
\usepackage{xcolor}
\usepackage{colortbl}
\newcolumntype{Y}{>{\raggedright\arraybackslash}X}

\title{Reasoning Steps as Curriculum:\\Using Depth of Thought as a Difficulty Signal for Tuning LLMs}

\author{
 \textbf{Jeesu Jung\textsuperscript{1}},
 \textbf{Sangkeun Jung\textsuperscript{1}}\thanks{Corresponding Author}
\\
 \textsuperscript{1} Chungnam National University, \\
  Department of Computer Science \& Engineering
\\
 \small{
 \texttt{jisu.jung5@gmail.com}, \texttt{hugmanskj@gmail.com}
 }
}

\begin{document}
\maketitle
\begin{abstract}
Curriculum learning for training LLMs requires a difficulty signal that aligns with reasoning while remaining scalable and interpretable. We propose a simple premise: tasks that demand deeper \emph{depth of thought} for humans should also be harder for models. Accordingly, we define difficulty as depth of thought (DoT) and \emph{operationalize} it by counting the discrete steps in a teacher model's reasoning trace (e.g., Chain-of-Thought). We then train with a shallow $\rightarrow$ deep curriculum ordered by this DoT and outline how to derive, validate, and schedule it at scale.
Our position yields three testable hypotheses: (i) DoT correlates with conventional difficulty on reasoning benchmarks, (ii) DoT-ordered curricula outperform length- or judge-scored curricula under matched budgets, and (iii) the difficulty is robust across teacher models given light formatting controls. We propose an evaluation framework and discuss threats to validity (teacher style, length confounds) alongside practical mitigations. Taken together, we aim to move toward cognitively grounded, interpretable curricula for reasoning-centric training.
\end{abstract}

\section{Introduction}
In recent years, large language models (LLMs) have been increasingly applied to advanced reasoning tasks, spanning high-school/college-level mathematics and science, multi-hop commonsense QA, and symbolic problem solving—where multi-step inference is essential~\cite{cot2022,long-important,iccl2024}. 

Motivated by this shift, alignment and training for LLMs increasingly target tasks that require multi-step reasoning rather than surface-level pattern matching~\cite{phased-ift,iccl2024}. In this setting, curriculum learning is appealing but hinges on a defensible notion of example difficulty that (i) targets reasoning ability, (ii) scales to millions of examples, and (iii) remains interpretable for humans who curate data and diagnose failures.

\paragraph{What should a difficulty signal satisfy?} A practical signal for reasoning-centric curricula should be: (1) \emph{structure-aware} (sensitive to the internal organization of reasoning rather than only output length), (2) \emph{model-robust} (not tightly coupled to a particular checkpoint or loss landscape), (3) \emph{computationally light} to compute at corpus scale, and (4) \emph{transparent} enough to explain data ordering decisions to practitioners.

Existing practices estimate difficulty with prompted judge scores or reward models, or rely on heuristics such as token length. These approaches can be effective but are often prompt- and domain-sensitive, can drift with model updates, and are only loosely aligned with the structure of reasoning.

\paragraph{Our premise: Depth of Thought (DoT).} We adopt a simple, testable premise: the deeper the required \emph{depth of thought} (DoT), the harder the example. We therefore \textbf{define difficulty as DoT}. To make DoT measurable at scale, we \emph{operationalize} it by counting discrete steps in a teacher model's reasoning trace (e.g., Chain-of-Thought, CoT) under consistent formatting. DoT aims to capture structural complexity beyond token length: multiple concise steps that chain intermediate claims differ meaningfully from a single verbose paragraph.

\begin{figure*}
    \centering
    \includegraphics[width=\linewidth]{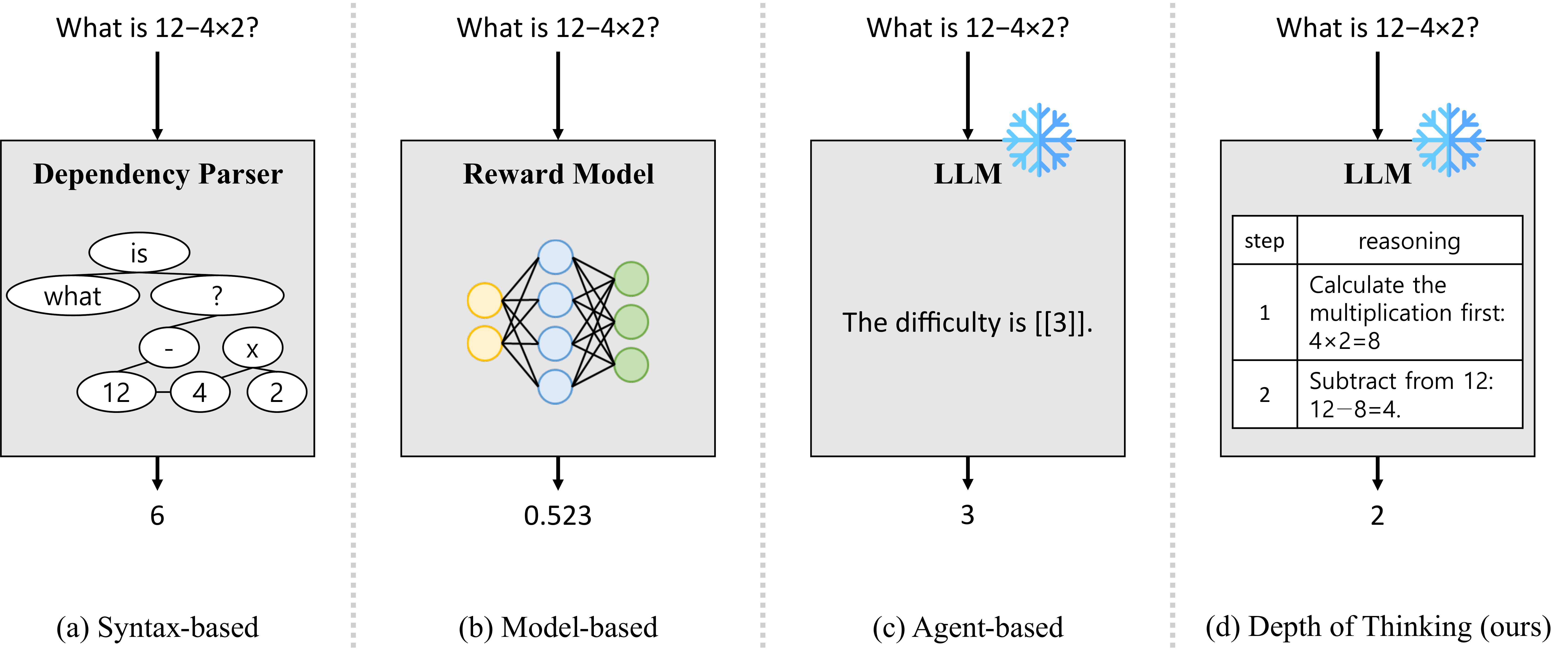}
    \caption{Comparison between conventional difficulty estimation methods and the proposed Depth of Thought-based approach.}
    \label{fig:Comparision}
\end{figure*}

\paragraph{Why this matters.} A DoT-ordered curriculum provides an interpretable basis for data scheduling (shallow $\rightarrow$ deep) that aligns with long-standing pedagogical intuition: arithmetic single-step recall precedes multi-step algebraic derivation; single-hop QA precedes multi-hop factual synthesis. Because DoT is a count derived from reasoning traces, curators can audit and adjust thresholds, and practitioners can attribute gains or failures to concrete changes in step distributions.

\paragraph{This paper's position and agenda.} We argue that DoT is a reasonable primary signal for reasoning curricula and outline a framework for deriving, validating, and scheduling it at scale. Concretely, we: (a) articulate desiderata for difficulty signals in reasoning-centric training; (b) propose DoT and its operationalization via reasoning traces; (c) compare DoT qualitatively to length-, judge-, and reward-based signals; and (d) present an evaluation framework and threats-to-validity analysis.

This paper advances three hypotheses: (H1) DoT correlates with conventional difficulty on reasoning benchmarks; (H2) DoT-ordered curricula outperform length-only or judge-scored curricula under matched budgets; (H3) the DoT signal is robust across teacher models when explicit numbering/separators are enforced.

\textbf{Scope.} We focus on student models supervised by teacher reasoning traces. The proposed signal complements, rather than replaces, existing curricula; we discuss limitations (teacher-style bias, step segmentation noise) and mitigations, and we describe how the agenda could be combined with other signals.

We next situate our work relative to prior art (Section~\ref{sec:related}), discuss limitations of current approaches (Section~\ref{sec:limits}), motivate DoT as difficulty (Section~\ref{sec:dot}), address counterarguments (Section~\ref{sec:counter}), and outline our methodology (Section~\ref{sec:method}).

\section{Related Work}\label{sec:related}
We group prior work along three axes: (i) how difficulty is quantified, (ii) where curriculum is applied (in-context, supervised fine-tuning, or RL), and (iii) how reasoning traces (e.g., CoT) are used (as supervision, signal, or metric). A simple comparative example of these categories can be seen in Figure~\ref{fig:Comparision}.

\subsection{Syntax-based Difficulty}
Early work often approximated difficulty using surface-form proxies such as token/sentence length, clause counts, and parse-tree depth because they are model-agnostic and inexpensive. However, these measures primarily capture verbosity and syntactic complexity rather than the \emph{structure} of reasoning. They can be inflated by redundant phrasing, vary with tokenizers and domains, and correlate only weakly with the multi-step dependencies required by hard reasoning problems~\cite{long-important}.

\subsection{Model-based Difficulty}
Model-centric proxies define difficulty via training dynamics—most commonly per-example loss and gradient magnitudes. While these measures are easy to obtain during training, they are inherently checkpoint- and architecture-dependent, susceptible to instability as optimization proceeds, and risk conflating model limitations with task-intrinsic complexity. Gradient-based variants further impose nontrivial computational cost when computed per example at scale.

\subsection{LLM-based Difficulty}
With recent advances in large language models, difficulty estimation has shifted toward LLM-aware measures. Two prevalent directions are (i) \emph{LLM-as-a-judge} scores, where a strong model assigns ratings to examples or outputs, and (ii) \emph{reward models} trained to predict scalar preferences or success signals. These measures can be more task-aware than syntax heuristics and can adapt to diverse domains, but they introduce new limitations: sensitivity to prompts and domains, dependence on the particular judge/reward model, and limited interpretability for data curation and audit~\cite{acl2024-naacl82,gpt-self-eval,teacher-eval}. Orthogonal evidence further suggests that length alone is an unreliable stand-in for reasoning structure~\cite{long-important}.

\subsection{Curriculum Learning for LLMs}
Curriculum ideas have been explored in LLMs across settings. In in-context learning (ICL), demonstrations are ordered from easy to hard to improve few-shot reasoning~\cite{iccl2024}. The most related ICL approach ranks examples by the number of problem-solving operations (a notion of step count) and presents them in a curriculum order~\cite{problem-solving-icl}; importantly, this operates at prompt time and does not change model parameters. For supervised fine-tuning (SFT), phased or staged curricula use external difficulty scores to partition data and train from easy to hard~\cite{phased-ift}. RL-based approaches similarly adapt sampling with evolving curricula. Across these lines, curricula typically rely on heuristic or opaque difficulty signals rather than model-internal structure such as CoT steps.

\subsection{Reasoning Traces and Depth of Thought}
Externalizing intermediate reasoning (e.g., via CoT prompting) substantially improves problem solving by exposing stepwise structure~\cite{cot2022}. Yet existing uses of CoT in training either supervise on final answers with CoT as auxiliary text or exploit CoT for selection, not as a \emph{metric} to define data difficulty. Moreover, step count and token length are correlated but distinct: multiple short steps may reflect different cognitive moves than one verbose paragraph. This motivates treating \emph{depth of thought} as an interpretable, objective proxy for difficulty that targets reasoning structure, and investigating its role in curriculum design.

\subsection{Summary}
Current approaches to curriculum learning in language models have significant limitations: \textbf{they often measure difficulty from the model's perspective rather than the task's intrinsic complexity}. Loss-based metrics are model-dependent and unstable during training. Gradient-based measures require expensive computation. Human annotations are subjective and unscalable.

\section{Current Approaches and Their Limitations}\label{sec:limits}

\subsection{Syntax-based Difficulty}
Heuristics based on surface syntax—e.g., token or sentence length, constituency/dependency tree depth, or clause counts—are attractive because they are model-agnostic and inexpensive. However, they often fail to separate truly hard reasoning problems from merely verbose ones. In practice, such measures can be gamed by adding redundant phrases, vary with tokenizer/domain shifts, and correlate only weakly with the structural demands of multi-step reasoning~\cite{long-important}. As a result, syntax-only measures tend to mis-rank examples whose difficulty stems from latent multi-hop dependencies rather than surface form.

\subsection{Model-based Difficulty: Loss and Gradient Proxies}
Using training loss as a difficulty measure creates a circular dependency: we measure difficulty by how hard the current checkpoint finds an example, then use that measure to decide what to train next. This conflates model limitations with task complexity and can be unstable as parameters, data order, or optimization settings change. Gradient-based variants require per-example forward--backward passes and are thus computationally heavy at scale; moreover, both loss and gradient magnitudes are checkpoint- and architecture-dependent, limiting portability across models and training stages.

\subsection{Agent-based Difficulty: LLM-as-a-Judge and Tool-Using Evaluators}
A popular alternative is to use an external agent to rate difficulty or quality—e.g., a strong LLM acting as a judge, possibly with tools—yielding scalar scores used for data ranking or selection. While these measures can be more task-aware than syntax heuristics, they introduce limitations: sensitivity to prompts and instructions, style biases and calibration drift across updates, dependence on the particular judge/reward setup, and nontrivial cost. They can also encourage reward hacking when models overfit to the judge's preferences rather than task structure. See findings on LLM-as-a-judge reliability and evaluator meta-evaluation for discussion~\cite{gpt-self-eval,teacher-eval,acl2024-naacl82}.

\subsection{Human-based Difficulty: Annotations and Expert Labels}
Human-judged difficulty suffers from inter-annotator disagreement and cannot scale to the millions of examples needed for effective LLM training. Moreover, human intuitions about difficulty may not align with what makes tasks challenging for neural networks.

\subsection{Positioning our proposal.} 
We seek a difficulty measure that remains scalable and model-robust while targeting reasoning \emph{structure}. Our approach operationalizes \emph{depth of thought} by counting discrete steps in reasoning traces (e.g., CoT), yielding an interpretable, task-oriented measure that complements and, in many cases, can replace opaque judge/reward scores~\cite{cot2022}.

\section{Depth of Thought as Difficulty}\label{sec:dot}

\begin{figure}
    \centering
    \includegraphics[width=\linewidth]{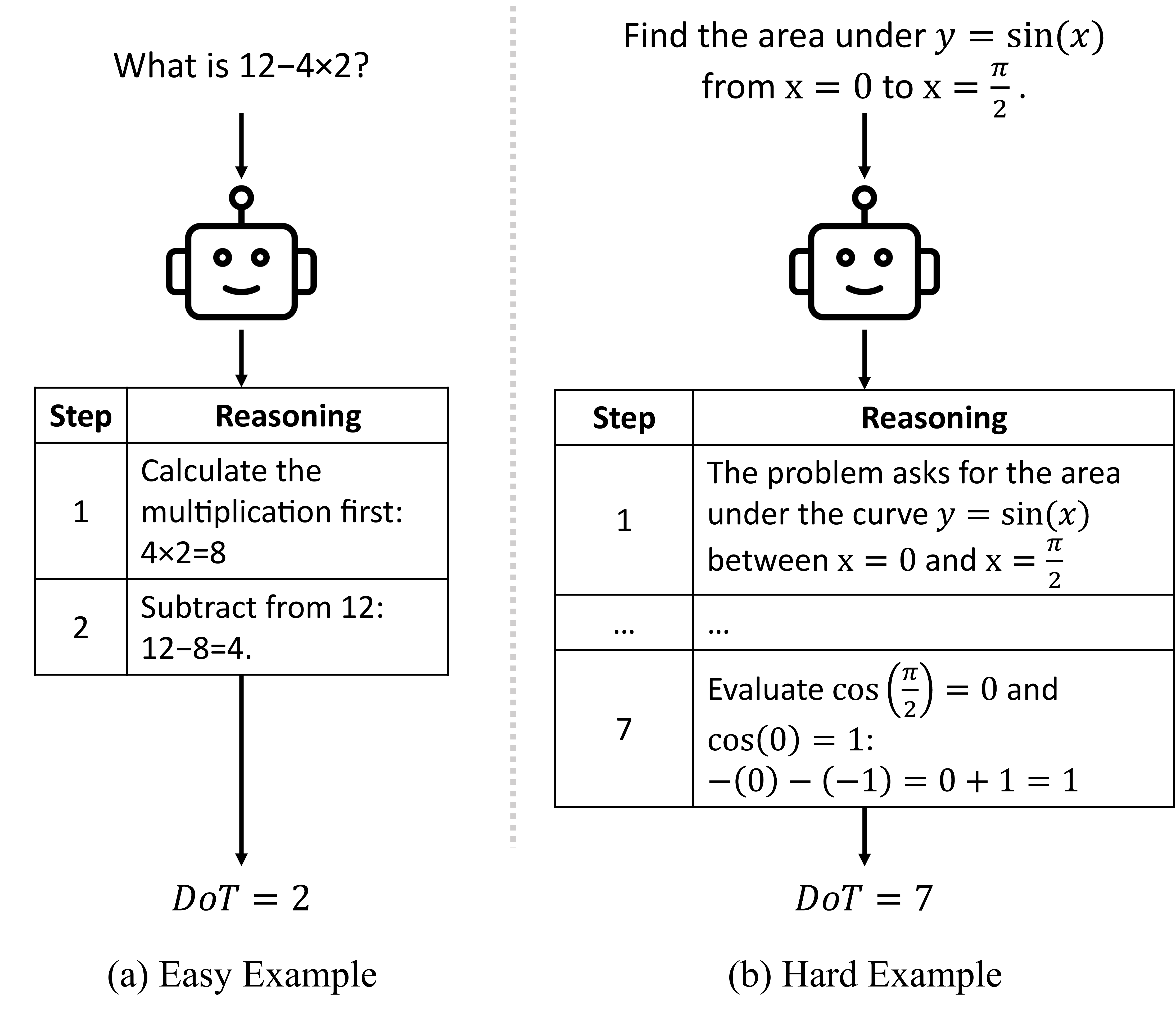}
    \caption{Illustration of difficulty computation using Depth of Thought (DoT). Even with similar input lengths, the reasoning process lengthens as difficulty increases.}
    \label{fig:difficulty}
\end{figure}
\subsection{Computational Depth as the Core of Cognitive Complexity}

\textbf{The depth of reasoning required to solve a problem directly corresponds to its inherent difficulty.} In human cognition, we instinctively recognize that problems requiring more intermediate steps are inherently more difficult. A single-step arithmetic calculation ($2+3$) is fundamentally easier than a multi-step algebraic equation requiring variable substitution, factorization, and simplification.

This correspondence between reasoning depth and difficulty is not merely correlational--it represents the \textbf{core definition of cognitive complexity}. When we say a problem is ``difficult,'' we primarily mean it requires more elaborate thought processes, more intermediate conclusions, and more careful logical progression. The number of reasoning steps captures this essential characteristic of difficulty in its purest form. 
A simple example illustrating this can be found in Figure~\ref{fig:difficulty}.

\subsection{Improved Interpretability and Explainability}

Current difficulty metrics provide little insight into \emph{why} a task is challenging or \emph{how} to structure learning progression. Loss-based metrics offer numerical values without semantic meaning. Gradient-based measures remain opaque to practitioners.

\textbf{Depth-based difficulty can make curriculum design more transparent and interpretable} by reducing reliance on black-box optimization.

Moreover, depth-based curricula can be easily communicated to stakeholders, validated by domain experts, and adapted across different subjects while maintaining conceptual coherence.

\begin{figure*}[ht]
    \centering
    \includegraphics[width=\linewidth]{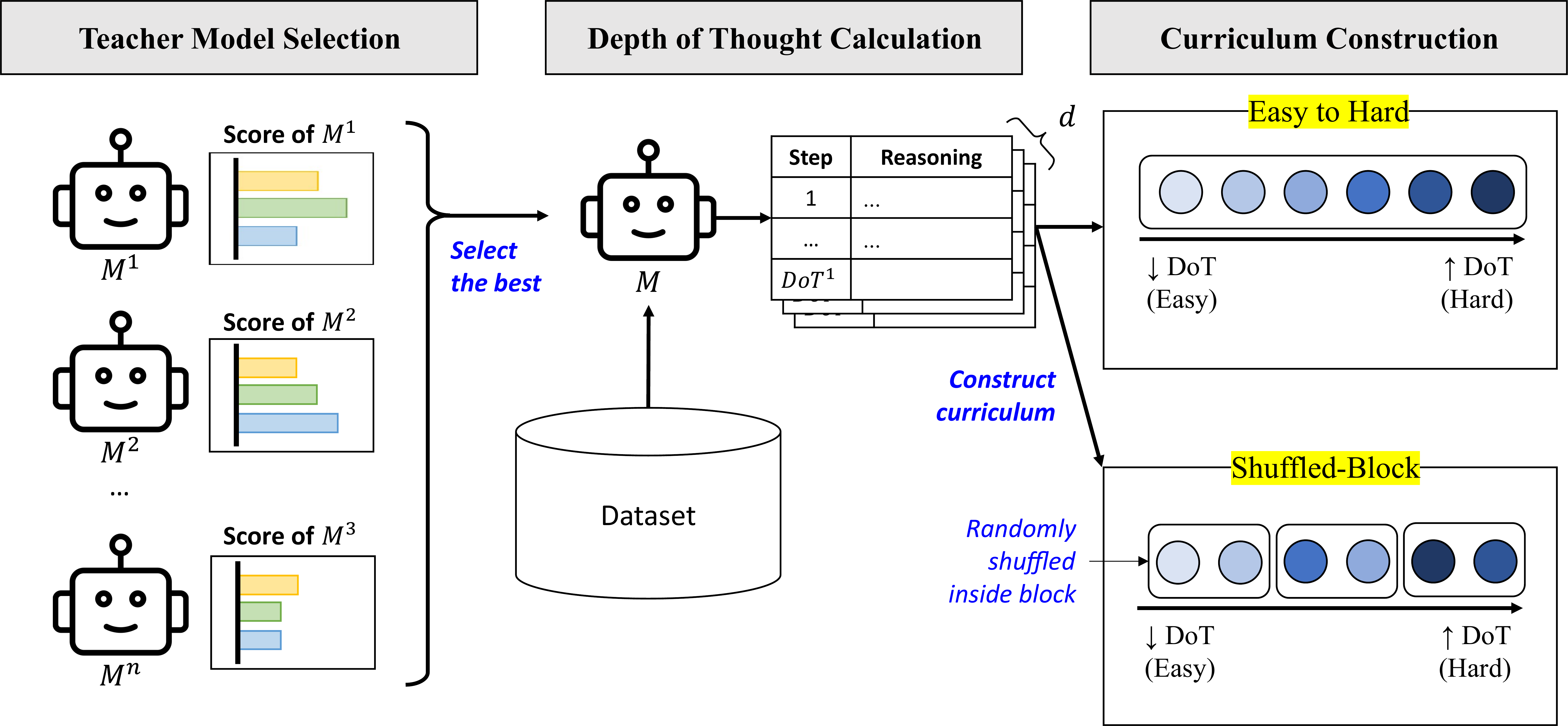}
    \caption{Overall framework for constructing a DoT-driven curriculum. The process includes teacher model selection, reasoning trace generation, and curriculum scheduling from easier (lower DoT) to harder (higher DoT) examples. $M^i$ is the candidate model, and $d$ is the size of the entire dataset.}
    \label{fig:framework}
\end{figure*}

\section{Addressing Potential Counterarguments}\label{sec:counter}

\subsection{Reasoning Depth Doesn't Capture All Aspects of Difficulty}

\textbf{Response}: We don't claim reasoning depth captures every dimension of difficulty, but we argue it captures the most important one for reasoning tasks. Other factors (domain knowledge, linguistic complexity) can be incorporated as secondary metrics, but reasoning complexity should be the primary curriculum driver.

\subsection{Generated Reasoning Steps May Be Unreliable or Inconsistent}

\textbf{Response}: This concern highlights the critical importance of rigorous teacher model selection and validation. We address this through systematic evaluation of candidate teacher models on reasoning benchmarks, selecting only those with demonstrated strong reasoning performance. Stronger reasoning models tend to generate more reliable stepwise traces, and the resulting difficulty assessments may generalize across datasets and tasks. Additionally, we may use stochastic self-consistency and prompt ensembling to improve stability and reliability of the measured difficulty.

\subsection{Simple Tasks May Have Unnecessarily Long Reasoning Chains}

\textbf{Response}: This represents a failure of reasoning generation, not our framework. A properly generated reasoning chain should reflect the minimal necessary reasoning steps. Overly verbose reasoning indicates poor prompt engineering or model instruction-following, not fundamental flaws in using reasoning depth as difficulty.

\section{Operational Framework: DoT Derivation and Curriculum Construction}\label{sec:method}

\subsection{Problem Definition}
In this work, we aim to define difficulty such that it (1) is agnostic to superficial syntactic structure, (2) can be scored in a human-like manner, (3) affords human-interpretable explanations, and (4) does not rely on additional training during knowledge transfer. 

To this end, we leverage the depth of thought of a high-capacity teacher model to extract step-level concepts from the actual reasoning process and use them to train the student model.

We formalize \emph{depth of thought} (DoT) for an example $x$ using the teacher's reasoning trace $c=[s_1,\dots,s_k]$ segmented into discrete steps. We define
\begin{equation}
\mathrm{DoT}(x)=k, \quad \mathrm{DoT}_{\mathrm{norm}}(x)=\frac{k}{\log(1+\mathrm{tok}(c))},
\end{equation}
where $\mathrm{tok}(c)$ is the token length of the trace. The normalized variant controls for verbosity while preserving step structure. We then construct a curriculum by ordering examples from shallow to deep DoT and scheduling training phases accordingly. In practice, we:

\begin{itemize}
    \item enforce explicit numbering/separators in teacher prompts to make step boundaries unambiguous~\cite{cot2022};
    \item parse steps with simple regex/markup rules plus spot-check audits; and
    \item bucket examples by DoT (e.g., 1--3, 4--6, 7+) with task balance per bucket.
\end{itemize}

We compare staged and mixed curricula by sampling from the union of buckets up to phase $t$, using weights $w_i\propto i^{\alpha}$ (sharpness $\alpha\ge0$), which recovers hard staging when $\alpha\to\infty$ and uniform mixing when $\alpha=0$~\cite{phased-ift,iccl2024}.

\subsection{Overall Process}

To build such a dataset, we proceed according to the following steps:
\begin{enumerate}
\item \textbf{Teacher Model Selection}: Systematically evaluate multiple state-of-the-art LLMs (GPT-4, Claude, Gemini, etc.) on reasoning benchmarks to identify the most capable teacher model for CoT generation.
\item \textbf{Reasoning-trace Collection (e.g., CoT)}: Generate high-quality CoT explanations using the selected teacher model across standardized reasoning datasets with explicit numbering templates.
\item \textbf{Curriculum Construction}: Create difficulty-graded curricula based on validated DoT counts, scheduling shallow $\rightarrow$ deep with optional mixing between adjacent buckets.
\item \textbf{Universal Evaluation}: Train diverse small LLM architectures with DoT-based curricula and compare against token-length and judge-scored curricula under matched budgets to assess breadth of applicability.
\end{enumerate}

The overall process of curriculum construction is shown in Figure~\ref{fig:framework}.

\section{Expected Benefits of DoT-based Difficulty}
DoT-driven curriculum can support several practical benefits.

\begin{itemize}
    \item \textbf{Beyond-length}: Move beyond reliance on token or surface syntactic complexity by computing difficulty from the actual depth of thought.
    \item \textbf{Annotation-free curriculum}: Enable human-like curricula without human annotation by training models sequentially according to DoT.
    \item \textbf{Interpretability \& control}: Provide interpretable, controllable scheduling—e.g., focus on easier data by training only on traces with $\leq$2 steps, or emphasize advanced reasoning by requiring traces with $\geq$5 steps.
    \item \textbf{Efficient transfer}: Facilitate knowledge transfer by conveying concepts to the student model with little or no additional task-specific training.
\end{itemize}

\section{Conclusion}

The AI community has focused too heavily on what models find difficult rather than what tasks are inherently difficult. Reasoning depth offers a path toward more principled, interpretable, and effective curriculum learning.

\textbf{Our position}: The reasoning complexity inherent in a problem, as reflected by the depth of thinking required, should drive curriculum design rather than model-dependent difficulty proxies. This cognitive-alignment principle will prove more effective for training reasoning-capable AI systems.

We call on the community to move beyond ad-hoc difficulty metrics toward cognitively grounded approaches that align AI training with human learning principles. The success of depth-based curriculum learning could fundamentally reshape how we approach AI education, making it more interpretable, effective, and aligned with human cognition.

\bibliographystyle{acl_natbib}
\bibliography{custom}

\end{document}